\documentclass[journal]{IEEEtran}
\usepackage{amssymb}
\usepackage{amsmath}
\usepackage{graphicx}
\usepackage{url}
\usepackage{array}
\usepackage{color,soul} % highlighting
\sethlcolor{white} % uncomment to switch highlights off
\soulregister\ref{7} % make soul work with references
\soulregister\cite{7} % make soul work with citations
\usepackage{float} % stop figure from floating with [htb] (package {here} is obsolete
\usepackage{subcaption}
\usepackage[colorinlistoftodos]{todonotes}
\usepackage{booktabs} % To thicken table lines
\usepackage{rotating}
\usepackage{multirow}
\usepackage{transparent}
\usepackage{color}
\usepackage{bm} % bold math symbols
\newcommand*{\vcenteredhbox}[1]{\begingroup\setbox0=\hbox{#1}\parbox{\wd0}{\box0}\endgroup}
\def \Npatients {82}
\def \Ntrain {$\sim$60}
\def \Ntest {$\sim$20}
\newcommand\figscale{0.95}

\begin{document}
\title{Spatial Aggregation of Holistically-Nested Convolutional Neural Networks for Automated Pancreas Localization and Segmentation} 
\author{Holger R. Roth, Le Lu, {\em Senior Member}, {\em IEEE}, Nathan Lay, Adam P. Harrison, Amal Farag, Andrew Sohn, Ronald M. Summers
\thanks{All authors are with the Imaging Biomarkers and Computer-Aided Diagnosis Laboratory, Clinical Image Processing Service, Radiology and Imaging Sciences Department, National Institutes of Health Clinical Center, Bethesda, MD 20892-1182, USA. This work was supported by the Intramural Research Program of National Institutes of Health Clinical Center. This version was submitted to IEEE Trans. on Medical Imaging on Dec. 18th, 2016. The content of this article is covered by US Patent Applications of 62/345,606\# and 62/450,681\#. Contact emails: \{le.lu, rms\}@nih.gov.
}% <-this % stops a space
} % \author
%%%%%%%%%%%%%%%%%%%%%%%%%%%%%%%%%%%%%%%%%%%%%%%%%%%%%%%%%%%%%%%%%%%%%%%%%%%%%%%%%%%%%%%
\maketitle
%%%%%%%%%%%%%%%%%%%%%%%%%%%%%%%%%%%%%%%%%%%%%%%%%%%%%%%%%%%%%%%%%%%%%%%%%%%%%%%%%%%%%%%
\begin{abstract}
Accurate and automatic organ segmentation from 3D radiological scans is an important yet challenging problem for medical image analysis. Specifically, as a small, soft, and flexible abdominal organ, the pancreas demonstrates very high inter-patient anatomical variability in both its shape and volume. This inhibits traditional automated segmentation methods from achieving high accuracies, especially compared to the performance obtained for other organs, such as the liver, heart or kidneys. 
%%%%%%%%%%%%%%%%%%%%%%%%%%%%%%%%%%%%%%%%%%%%%%%%%%%%%%%%
To fill this gap, we present an automated system from 3D computed tomography (CT) volumes that is based on a two-stage cascaded approach---pancreas localization and pancreas segmentation. For the first step, we localize the pancreas from the entire 3D CT scan, providing a reliable bounding box for the more refined segmentation step. We introduce a fully deep-learning approach, based on an efficient application of holistically-nested convolutional networks (HNNs) on the three orthogonal axial, sagittal, and coronal views. The resulting HNN per-pixel probability maps are then fused using pooling to reliably produce a 3D bounding box of the pancreas that maximizes the recall. We show that our introduced localizer \hl{compares favorably} to both a conventional non-deep-learning method and a recent hybrid approach based on spatial aggregation of superpixels using random forest classification. The second, segmentation, phase operates within the computed bounding box and integrates semantic mid-level cues of deeply-learned organ {\em interior} and {\em boundary} maps, obtained by two additional and separate realizations of HNNs. By integrating these two mid-level cues, our method is capable of generating boundary-preserving pixel-wise class label maps that result in the final pancreas segmentation. 
%%%%%%%%%%%%%%%%%%%%%%%%%%%%%%%%%%%%%%%%%%%%%%%%%%%%%%%%
Quantitative evaluation is performed on a publicly available dataset of \Npatients{} patient CT scans using 4-fold cross-validation (CV). We achieve a (mean $\pm$ std. dev.) Dice similarity coefficient (DSC) of 81.27$\pm$6.27\% in validation, which significantly outperforms both a previous state-of-the art method and a preliminary version of this work that report DSCs of 71.80$\pm$10.70\% and 78.01$\pm$8.20\%, respectively, using the same dataset.
\end{abstract}
%%%%%%%%%%%%%%%%%%%%%%%%%%%%%%%%%%%%%%%%%%%%%%%%%%%%%%%%%%%%%
%%%%%%%%%%%%%%%%%%%%%%%%%%%%%%%%%%%%%%%%%%%%%%%%%%%%%%%%%%%%%
\IEEEpeerreviewmaketitle
\section{Introduction}
\IEEEPARstart{P}{ancreas} segmentation in computed tomography (CT) challenges current computer-aided diagnosis (CAD) systems. While automatic segmentation of numerous other organs in CT scans, such as the liver, heart or kidneys, achieves good performance with Dice similarity coefficients (DSCs) of $>$90\% \cite{Wang2014Miccai,Chu2013Miccai,wolz2013automated}, the pancreas' variable shape, size, and location in the abdomen limits segmentation accuracy to $<$73\% DSC being reported in the literature \cite{wolz2013automated,Chu2013Miccai,tong2015discriminative,okada2015abdominal,farag2014bottom,roth2015deeporgan}. Previous pancreas segmentation work \cite{wolz2013automated,Chu2013Miccai,tong2015discriminative,okada2015abdominal} are all based on performing volumetric multiple atlas registration \cite{Modat2010,avants2009advanced,Avants2011reproducible} and executing robust label fusion methods \cite{Wang2013Multi,Bai2013probabilistic,Wang2014Segmentation} to optimize the per-pixel organ labeling process. This type of organ segmentation strategy is widely used for many organ segmentation problems, such as the brain \cite{Wang2013Multi,Wang2014Segmentation}, heart \cite{Bai2013probabilistic}, lung \cite{Murphy2011Evaluation}, and pancreas \cite{wolz2013automated,Chu2013Miccai,tong2015discriminative,okada2015abdominal}. These methods can be referred as a {\em top-down} model fitting approach, or more specifically, MALF (Multi-Atlas Registration \& Label Fusion). Another group of top-down frameworks \cite{Ecabert2008Automatic,Zheng08,Ling08} leverages statistical model detection, e.g., generalized Hough transform \cite{Ecabert2008Automatic} or marginal space learning \cite{Zheng08,Ling08}, for organ localization; and deformable statistical shape models for object segmentation. However, due to the intrinsic huge 3D shape variability of the pancreas, statistical shape modeling has not been applied for pancreas segmentation.

Recently, a new {\em bottom-up} pancreas segmentation representation has been proposed in \cite{farag2014bottom}, which uses dense binary image patch labeling confidence maps that are aggregated to classify image regions, or superpixels \cite{Felzenszwalb2004Efficient,pont-tuset2015mcg,Girshick2016Region}, into pancreas and non-pancreas label assignments. This method's motivation is to improve segmentation accuracy of highly deformable organs, such as the pancreas, by leveraging mid-level visual representations of image segments. This work was advanced further by Roth et al.\cite{roth2015deeporgan}, who proposed a probabilistic bottom-up approach using a set of multi-scale and multi-level deep convolutional neural networks (CNNs) to capture the complexity of pancreas appearance in CT images. The resulting system improved upon the performance of \cite{farag2014bottom} with a reported DSC of 71.8$\pm$10.7\% against 68.8$\pm$25.6\%. Compared to the MALF based pancreas segmentation work \cite{wolz2013automated,Chu2013Miccai,tong2015discriminative,okada2015abdominal} that are evaluated using ``leave-one-patient-out'' (LOO) protocol, the bottom-up approaches using superpixel representation \cite{farag2014bottom,roth2015deeporgan} have reported comparable or higher DSC accuracy measurements, under more challenging 6-fold or 4-fold cross-validation\footnote{As discussed in \cite{Shin2016Deep}, LOO can be considered as an extreme case of $M$-fold cross-validation with $M=N$ when $N$ patient datasets are available for experiments. When $M$ is decreasing and significantly smaller than $N$, $M$-fold CV becomes more challenging since there are less data for training and more patient cases on testing.}. Comparing the two bottom-up approaches, the usage of deep CNN models has noticeably improved the performance stability, which is evident by the significantly smaller standard deviation \cite{roth2015deeporgan} than all other top-down or bottom-up works \cite{farag2014bottom,wolz2013automated,Chu2013Miccai,tong2015discriminative,okada2015abdominal}.  

Deep CNNs have successfully been applied to many high-level tasks in medical imaging, such as recognition and object detection \cite{yan2015bodypart}. The main advantage of CNNs comes from the fact that end-to-end learning of salient feature representations for the task at hand is more effective than hand-crafted features with heuristically tuned parameters \cite{zheng2015conditional}. Similarly, CNNs demonstrate promising performance for pixel-level labeling problems, e.g., semantic segmentation in recent computer vision and medical imaging analysis work, e.g., fully convolutional neural networks (FCN) \cite{long2015fully}, DeepLab \cite{chen2014semantic} and U-Net \cite{ronneberger2015unet}. These approaches have all garnered significant improvements in performance over previous methods by applying state-of-the-art CNN-based image classifiers and representation to the semantic segmentation problem in both domains. 

Semantic organ segmentation involves assigning a label to each pixel in the image. On one hand, features for classification of single pixels (or patches) play a major role, but on the other hand, factors such as edges, i.e., organ boundaries, appearance consistency, and spatial consistency, could greatly impact the overall system performance \cite{zheng2015conditional}. Furthermore, there are indications of semantic vision tasks requiring hierarchical levels of visual perception and abstraction \cite{xie2015holistically}. As such, generating rich feature hierarchies for both the interior and the boundary of the organ could provide important ``mid-level visual cues'' for semantic segmentation. Subsequent spatial aggregation of these mid-level cues then has the prospect of improving semantic segmentation methods by enhancing the accuracy and consistency of pixel-level labeling.

A preliminary version of this work appears as \cite{roth2016spatial}, where we demonstrate that a two-stage bottom-up localization and segmentation approach can improve upon the state of the art. In this work, the major extension is that we describe an improved pancreas localization method by replacing the initial super-pixel based one, with a new general deep learning based approach. This methodological component is designed to optimize or maximize the pancreas spatial recall criterion while reducing the non-pancreas volume as much as possible. Specifically, we generate the per-pixel pancreas class probability maps (or ``heat maps'') through an efficient combination of holistically-nested convolutional networks (HNNs) in the three orthogonal axial, sagittal, and coronal CT views. We fuse the three HNN outputs to produce a 3D bounding box covering the underlying, yet latent in testing, pancreas volume by nearly 100\%. In addition, we show that exactly the same HNN model architecture can be effective for the subsequent pancreas segmentation stage by integrating both deeply learned boundary and appearance cues. This also results in a simpler overall pancreas localization and segmentation system using HNNs only, rather than the previous hybrid setup involving non-deep- and deep-learning method components\cite{roth2016spatial}. Lastly, our current method reports an overall improved DSC performance compared to \cite{roth2016spatial} and \cite{roth2015deeporgan}: DSC of 81.14$\pm$7.3\% versus 78.0$\pm$8.2\% and 71.8$\pm$10.7\% \cite{roth2015deeporgan}, respectively.   

The proposed two-stage process essentially performs {\em 3D spatial aggregation and assembling} on the HNN-produced per-pixel pancreas probability maps that run on 2D axial, coronal, and sagittal CT planes. This process operates exhaustively for pancreas localization and selectively for pancreas segmentation. Therefore, this work inherits a hierarchical and compositional visual representation of computing 3D object information aggregated from 2D image slices or parts, in a similar spirit of \cite{roth2014new,Farabet2013Learning,Lu2008Accurate}. Alternatively, there are recent studies on directly using 3D convolutional neural networks for liver, brain segmentation \cite{Dou20163D,Chen2016VoxResNet} and volumetric vascular boundary detection \cite{Merkow2016Dense}. Due to CNN memory restrictions, these 3D CNN approaches adopt padded sliding windows or volumes to process the original CT scans, such as 96$\times$96$\times$48 segments \cite{Merkow2016Dense}, 160$\times$160$\times$72 subvolumes \cite{Dou20163D} and 80$\times$80$\times$80 windows \cite{Chen2016VoxResNet}, which may cause  segmentation discontinuities or inconsistencies at overlapped window boundaries. We argue that learning shareable lower-dimensional 2D CNN models may be more generalizable and handle the ``curse-of-dimensionality'' issue better than their fully 3D counterparts, especially when used to parse complex 3D anatomical structures, e.g., lymph node clusters \cite{Nogues2016LNC,roth2016improving} and the pancreas \cite{roth2015deeporgan,roth2016spatial}. 
%%%%%%%%%%%%%%%%%%%%%%%%%%%%%%%%%%%%%%%%%%%%%%%%%%%%%%%%
Analogous examples of comparing compositional multi-view 2D CNNs versus direct 3D deep models can be found in other computer vision problems: 1) video based action recognition where a two-stream 2D CNN model \cite{Simonyan2014Two}, capturing the image intensity and motion cues, significantly improves upon the 3D CNN method \cite{Karpathy2014Large}; 2) the advantageous performance of multi-view CNNs over volumetric CNNs in 3D Shape Recognition \cite{Su2015Multi}. 
%%%%%%%%%%%%%%%%%%%%%%%%%%%
\hl{The rest of this paper is organized as follows. We describe the technical motivation and details of the proposed approach in Sec. \ref{sec:methods}. Experimental results and comparison with related work are addressed in Sec. \ref{sec:results}. We conclude the paper, an with extended discussion, in Sec. \ref{sec:discussion}}.
%%%%%%%%%%%%%%%%%%%%%%%%%%%%%%%%%%%%%%%%%%%%%%%%%%%%%%%%%%%%%
%%%%%%%%%%%%%%%%%%%%%%%%%%%
%%%%%%%%%%%%%%%%%%%%%%%%%%%%%%%%%%%%%%%%%%%%%%%%%%%%%%%%%%%%%
\section{Methods}\label{sec:methods}
\begin{figure}[htb]%[htb]
\centering	
    \resizebox{0.9\columnwidth}{!}{\includegraphics{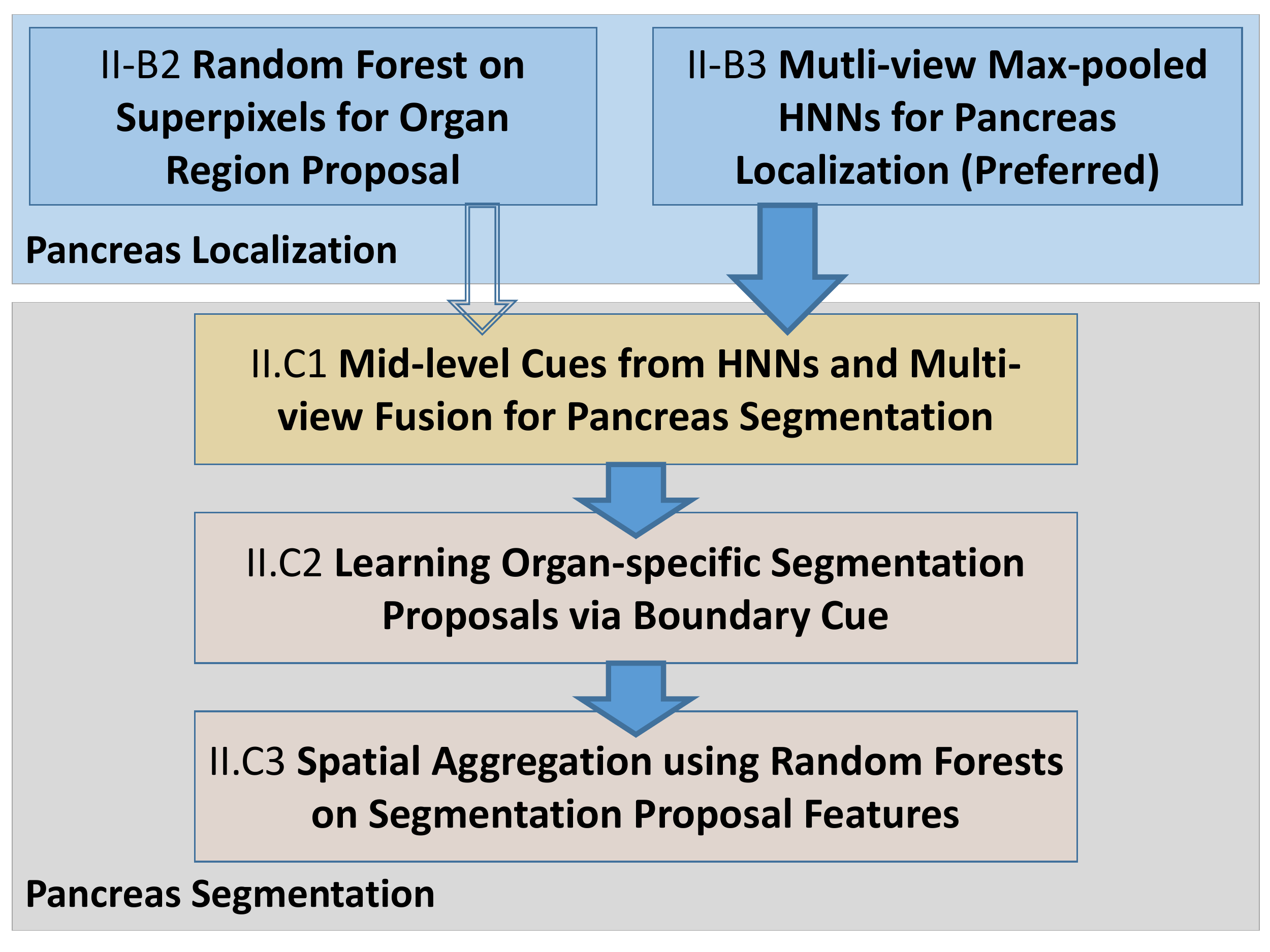}}
	\caption{\small Flowchart of the proposed two-stage pancreas localization and segmentation framework. Sec. \ref{sec:region_candidates_RF} and Sec. \ref{sec:localizationHNN} are the alternative means of bottom-up organ localization. The remaining modules are for pancreas segmentation.}
	\label{fig:flowchart}
\end{figure}

%%%%%%%%%%%%%%%%%%%%%%%%%%%
In this work, we present a two-phased approach for automated pancreas \textbf{localization} and \textbf{segmentation}. The pancreas localization step aims to robustly compute a bounding box which, at the desirable setting, should cover the entire pancreas while pruning the high majority volumetric space from any input CT scan without any manual pre-processing. The second stage of pancreas segmentation incorporates deeply learned organ interior and boundary mid-level cues with subsequent spatial aggregation, focusing only on the properly zoomed or cascaded pancreas location and spatial extents that are generated after the first phase. 
%%%%%%%%%%%%%%%%%%%%%%%%%%%
In Sec. \ref{sec:segHNN} we introduce the HNN model that proves effective for both stages. Afterwards, we focus on localization in Sec. \ref{sec:localization}, which discusses and contrasts a conventional approach to localization with newer CNN-based ones---a hybrid and a fully deep-learning approach. We show how the latter approach, which relies on HNNs, provides a simple, yet state-of-the-art, localization method. Importantly, it relies on the same HNN architecture as the later segmentation step. With localization discussed, we explain our segmentation approach in Sec. \ref{sec:segmentation}, which relies on combining semantic mid-level cues produced from HNNs. Our approach to organ segmentation is based on simple, reproducible, yet effective, machine-learning principles. In particular, we demonstrate the most effective configuration of our system is simply composed of cascading and aggregating outputs from six HNNs trained at three orthogonal views and two spatial scales. No multi-atlas registration or multi-label fusion techniques are employed. Fig.~\ref{fig:flowchart} provides a flowchart depicting the makeup of our system.
%%%%%%%%%%%%%%%%%%%%%%%%%%%
%%%%%%%%%%%%%%%%%%%%%%%%%%%
 %Our methods are evaluated on CT scans of \Npatients{} patients in 4-fold cross-validation (instead of ``leave-one-patient-out'' evaluation often used in other work \cite{Wang2014Miccai,Chu2013Miccai,wolz2013automated}). %Holistically-nested neural networks \cite{xie2015holistically}, as explained in Section \ref{sec:segHNN}, can be adopted in our work to produce the dense pixel-to-pixel organ segmentation confidence or probability maps, by a single feedforward path.
%%%%%%%%%%%%%%%%%%%%%%%%%%%%%%%%%%%%%%%%%%%%%%%%%%%%%%%%%%%%%
\subsection{Learning Mid-level Cues via Holistically-Nested Networks for Localization and Segmentation} \label{sec:segHNN} 
\begin{figure}[t]%[htb]
 \centering	
    \def\svgwidth{\figscale\columnwidth}
    \input{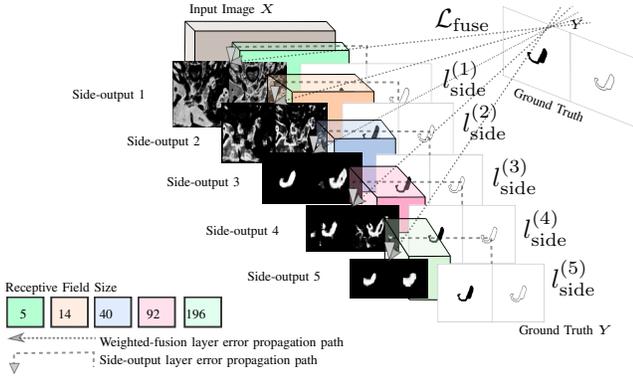}  
	\caption{\small The \textbf{HNN-I/B} network architecture for both interior (left images) and boundary (right images) detection pathways. We highlight the error back-propagation paths to illustrate the deep supervision performed at each side-output layer after the corresponding convolutional layer. As the side-outputs become smaller, the receptive field sizes get larger. This allows HNN to combine multi-scale and multi-level outputs in a learned weighted fusion layer. \hl{The ground truth images are inverted for aided visualization} (Figures adapted from \cite{xie2015holistically} with permission).}
	\label{fig:hed}
\end{figure}

%%%%%%%%%%%%%%%%%%%%%%%%%%%
\noindent In this work, we use the HNN architecture, to learn the pancreas' interior and boundary image-labeling maps, for both localization and segmentation. Object-level interior and boundary information are referred to as mid-level visual cues. Note that this type of CNN architecture was first proposed by \cite{xie2015holistically} under the name ``holistically-nested edge detection'' as a deep learning based general image edge detection method. %\hl{HNN trained on organ interior maps can effectively can be used for localization as described in Sec. \ref{sec:localizationHNN}.} 
It has been used successfully for extracting ``edge-like'' structures like blood vessels in 2D retina images \cite{fu2016retinal}. We however would argue and validate that it can serve as a suitable deep representation to learn general raw pixel-in and label-out mapping functions, i.e., to perform semantic segmentation. We use these principles to segment the interior of organs. 
%%%%%%%%%%%%%%%%%%%%%%%%%%%
HNN is designed to address two important issues: (1) training and prediction on the whole image end-to-end, i.e, holistically, using a per-pixel labeling cost; and (2) incorporating multi-scale and multi-level learning of deep image features \cite{xie2015holistically} via auxiliary cost functions at each convolutional layer. HNN computes the image-to-image or pixel-to-pixel prediction maps from any input raw image to its annotated labeling map, building on fully convolutional neural networks \cite{long2015fully} and deeply-supervised nets \cite{lee2014deeply}. The per-pixel labeling cost function \cite{long2015fully,xie2015holistically} makes it feasible that HNN/FCN can be effectively trained using only several hundred annotated image pairs. This enables the automatic learning of rich hierarchical feature representations and contexts that are critical to resolve spatial ambiguity in the segmentation of organs. The network structure is initialized based on an ImageNet pre-trained VGGNet model \cite{simonyan2014very}. It has been shown that fine-tuning CNNs pre-trained on general image classification tasks is helpful for low-level tasks, e.g., edge detection \cite{xie2015holistically}. Furthermore, we can utilize pre-trained edge-detection networks (trained on BSDS500 \cite{xie2015holistically}) to segment organ-specific boundaries.
%%%%%%%%%%%%%%%%%%%%%%%%%%%%%%%%%%%%%%%%%%%%%%%%%%%%%%%%%%%%%
%%%%%%%%%%%%%%%%%%%%%%%%%%%
\textbf{\textit{Network formulation:}} Our training data $S^{I/B} = \left\{(X_n , Y^{I/B}_n ), n = 1, \dots, N \right\}$ where $X_n$ denotes cropped axial CT images $X_n$, rescaled to within $\left[0,\dots,255\right]$ with a soft-tissue window of $[-160, 240]$ HU. $Y^I_n \in \left\{0,1\right\}$ and $Y^B_n \in \left\{0,1\right\}$ denote the binary ground truths of the interior and boundary map of the pancreas, respectively, for any corresponding $X_n$. Each image is considered holistically and independently as in \cite{xie2015holistically}. The network is able to learn features from these images alone from which interior  and boundary prediction maps can be produced, which we denote as \textbf{HNN-I} and \textbf{HNN-B}, respectively.
%%%%%%%%%%%%%%%%%%%%%%%%%%%
HNN can efficiently generate multi-level image features due to its deep architecture. Furthermore, multiple stages with different convolutional strides can capture the inherent scales of organ edge/interior labeling maps. However, due to the difficulty of learning such deep neural networks with multiple stages from scratch, we use the pre-trained network provided by \cite{xie2015holistically} and fine-tuned to our specific training data sets $S^{I/B}$ with a relatively smaller learning rate of $10^{-6}$. We use the HNN network architecture with 5 stages, including strides of 1, 2, 4, 8 and 16, respectively, and with different receptive field sizes as suggested by the authors\footnote{\scriptsize\url{https://github.com/s9xie/hed}.}.
%%%%%%%%%%%%%%%%%%%%%%%%%%%
In addition to standard CNN layers, a HNN network has $M$ side-output layers as shown in Fig. \ref{fig:hed}. These side-output layers are also realized as classifiers in which the corresponding weights are $\bm{\mathrm{w}} = (\bm{\mathrm{w}}^{(1)},\dots ,\bm{\mathrm{w}}^{(M )})$. For simplicity, all standard network layer parameters are denoted as $\bm{\mathrm{W}}$. Hence, the following objective function can be defined\footnote{\scriptsize We follow the notation of \cite{xie2015holistically}.}:
\begin{equation}
	\mathcal{L}_\mathrm{side}(\bm{\mathrm{W}},\bm{\mathrm{w}}) = \sum^{M}_{m=1}{\alpha_m}{l_\mathrm{side}^{(m)}}(\bm{\mathrm{W}},\bm{\mathrm{w}}^{m}).
	\label{equ:hed_loss}
\end{equation}
Here, $l_\mathrm{side}$ denotes an image-level loss function for side-outputs, computed over all pixels in a training image pair $X$ and $Y$. Because of the heavy bias towards non-labeled pixels in the ground truth data, \cite{xie2015holistically} introduces a strategy to automatically balance the loss between positive and negative classes via a per-pixel class-balancing weight $\beta$. This offsets the imbalances between edge/interior ($y=1$) and non-edge/exterior ($y=0$) samples. Specifically, a class-balanced cross-entropy loss function can be used in Eq. \eqref{equ:hed_loss} with $j$ iterating over the spatial dimensions of the image:
\begin{multline}
	l^{(m)}_\mathrm{side}(\bm{\mathrm{W}}, \bm{\mathrm{w}}^{(m)}) = - \beta \sum_{j\in Y_+}\log Pr\left(y_j=1|X;\bm{\mathrm{W}},\bm{\mathrm{w}}^{(m)}\right) - \\
	(1-\beta)\sum_{j\in Y_-}\log Pr\left(y_j=0|X;\bm{\mathrm{W}},\bm{\mathrm{w}}^{(m)}\right).
	\label{equ:hed_loss_balanced}
\end{multline}
Here, $\beta$ is simply $|Y_-|/|Y|$ and $1-\beta = |Y_+|/|Y|$, where $|Y_-|$ and $|Y_+|$ denote the ground truth set of \textit{negatives} and \textit{positives}, respectively. In contrast to \cite{xie2015holistically}, where $\beta$ is computed for each training image independently, we use a constant balancing weight computed on the entire training set. This is because some training slices might have no positives at all and otherwise would be ignored in the loss function. The class probability $Pr(y_j=1|X;W,w^{(m)}) = \sigma(a_j^{(m)}) \in [0,1]$ is computed on the activation value at each pixel $j$ using the sigmoid function $\sigma(.)$. Now, organ edge/interior map predictions $\hat{Y}ˆ{(m)}_\mathrm{side} = \sigma(\hat{A}ˆ{(m)}_\mathrm{side})$ can be obtained at each side-output layer, where $\hat{A}ˆ{(m)}_\mathrm{side} \equiv\{a_j^{(m)}, j = 1,\dots,|Y|\}$ are activations of the side-output of layer $m$. Finally, a ``weighted-fusion'' layer is added to the network that can be simultaneously learned during training. The loss function at the fusion layer $L_\mathrm{fuse}$ is defined as
\begin{equation}
	\mathcal{L}_\mathrm{fuse}(\bm{\mathrm{W}}, \bm{\mathrm{w}}, \bm{\mathrm{h}}) = Dist\left(Y, \hat{Y}_\mathrm{fuse}\right), 
\end{equation}
where $\hat{Y}_\mathrm{fuse} = \sigma \left(\sum^M_{m=1} h_m^{\hat{A}_\mathrm{side}}   \right)$ with $h = \left(h_1,\dots,h_M\right)$ being the fusion weight. $Dist(.,.)$ is a distance measure between the fused predictions and the ground truth label map. We use cross-entropy loss for this purpose. Hence, the following objective function can be minimized via standard stochastic gradient descent and back propagation:
\begin{equation}
	(\bm{\mathrm{W}}, \bm{\mathrm{w}}, \bm{\mathrm{h}})^\star = \mathrm{argmin}\left(\mathcal{L}_\mathrm{side}(\bm{\mathrm{W}}, \bm{\mathrm{w}}) + \mathcal{L}_\mathrm{fuse}(\bm{\mathrm{W}}, \bm{\mathrm{w}}, \bm{\mathrm{h}})\right)
\end{equation}
 %%%%%%%%%%%%%%%%%%%%%%%%%%%
\textit{Testing phase:} Given image $X$, we obtain both interior (\textbf{HNN-I}) and boundary (\textbf{HNN-B}) predictions from the models' side output layers and the weighted-fusion layer as in \cite{xie2015holistically}:
\begin{equation}
	\left(\hat{Y}^{I}_\mathrm{fuse}, \hat{Y}_\mathrm{side}^{I_1)}, \dots, \hat{Y}_\mathrm{side}^{I_M}\right) =  \textbf{HNN-I}\left(X, (\bm{\mathrm{W}}, \bm{\mathrm{w}}, \bm{\mathrm{h}}) \right)
\end{equation}
\begin{equation}
	\left(\hat{Y}^{B}_\mathrm{fuse}, \hat{Y}_\mathrm{side}^{B_1)}, \dots, \hat{Y}_\mathrm{side}^{B_M}\right) =  \textbf{HNN-B}\left(X, (\bm{\mathrm{W}}, \bm{\mathrm{w}}, \bm{\mathrm{h}}) \right) 
\end{equation}
Here, $\textbf{HNN-I/B}(\cdot)$ denotes the interior/boundary prediction maps estimated by the network. 
%%%%%%%%%%%%%%%%%%%%%%%%%%%%%%%%%%%%%%%%%%%%%%%%%%%%%%%%%%%%%%
%%%%%%%%%%%%%%%%%%%%%%%%%%%%%%%%%%%%%%%%%%%%%%%%%%%%%%%%%%%%%
%%%%%%%%%%%%%%%%%%%%%%%%%%%%%%%%%%%%%%%%%%%%%%%%%%%%%%%%%%%%%
%%%%%%%%%%%%%%%%%%%%%%%%%%%
\subsection{Pancreas Localization} \label{sec:localization}
Segmentation performance can be enhanced if irrelevant regions of the CT volume are pruned out. Conventional organ localization methods using random forest regression \cite{criminisi2013regression,laybirkbeck2013}, which we explain in Sec. \ref{sec:regression}, may not guarantee that the regressed organ bounding box contains the targeted organ with extremely high sensitivities on the pixel-level coverage. In Sec. \ref{sec:region_candidates_RF} we outline a superpixel based approach \cite{farag2014bottom}, based on hand-crafted and CNN features, that is able to provide improved performance. While this is effective, the complexity involved motivates our own development of a simpler and more accessible newly proposed multi-view HNN fusion based procedure. This is explained in Sec. \ref{sec:localizationHNN}. The output of the localization method will later feed into a more detailed and accurate segmentation method combining multiple mid-level cues from HNNs as illustrated in Fig. \ref{fig:flowchart}.
%%%%%%%%%%%%%%%%%%%%%%%%%%%%%%%%%%%%%%%%%%%%%%%%%%%%%%%%%%%%%%
%%%%%%%%%%%%%%%%%%%%%%%%%%%
%%%%%%%%%%%%%%%%%%%%%%%%%%%%%%%%%%%%%%%%%%%%%%%%%
\subsubsection{Regression Forest}\label{sec:regression}
Object localization by regression has been studied extensively in the literature including~\cite{criminisi2013regression,regressionKidneys,laybirkbeck2013}. The general idea is to predict an offset vector $\Delta x \in \mathbb{R}^3$ for a given image patch $I(x)$ centered about $x \in \mathbb{R}^3$. The predicted object position is then given as $x + \Delta x$. This is repeated for many examples of image patches and then aggregated to produce a final predicted position. Aggregation can be done with non-maximum suppression on prediction voting maps, mean aggregation~\cite{criminisi2013regression}, \hl{cluster medoid} aggregation~\cite{regressionKidneys}, and the use of local appearance with discriminative models to accept or reject predictions~\cite{laybirkbeck2013}.
%%%%%%%%%%%%%%%%%%%%%%%%%%%
The pancreas can be localized by regression due to their locations in the body in correlation to other anatomical structures. The objective is to predict bounding boxes $(x_{\text{center}},\Delta x_{\text{lower}}, \Delta x_{\text{upper}}) \in \mathbb{R}^{3 \times 3}$ where $x_{\text{center}}$ is the center of the pancreas and $x_{\text{center}} + \Delta x_{\text{lower}}$ and $x_{\text{center}} + \Delta x_{\text{upper}}$ are the lower and upper corner of the pancreas bounding box respectively. The addition of the extra three parameters follows from the observation that the center of the bounding box is not necessarily the center of the localized object. The pancreas Regression Forest predicts $(\Delta x, \Delta x_{\text{lower}}, \Delta x_{\text{upper}})$ for a given image patch $I(x)$. This produces pancreas bounding box candidates of the form $(x + \Delta x, \Delta x_{\text{lower}}, \Delta x_{\text{upper}})$. We additionally use a discriminative model to accept or reject predictions $x + \Delta x$ as in~\cite{laybirkbeck2013}. Finally, accepted predictions are aggregated using non-maximum suppression over probability scores and then the bounding boxes are ranked by the count of accepted predictions within the box. The box with the highest count of predictions is kept as the final prediction. %Performance of the system for pancreas detection is presented in Table~\ref{tbl:PancreasForestPerformance}.
\subsubsection{Random Forest on Superpixels}
\label{sec:region_candidates_RF}
As a form of initialization, we alternatively employ a previously proposed method based on random forest (RF) classification \cite{farag2014bottom,roth2015deeporgan} using both hand-crafted and deep CNN derived image features to compute a candidate bounding box regions. We only operate the RF labeling at a low probability threshold of $>$0.5 which is sufficient to reject the vast amount of non-pancreas from the CT images. This initial candidate generation is sufficient to extract bounding box regions that nearly surround the pancreases completely in all patient cases with $\sim97$\% recall. All candidate regions are computed during the testing phase of cross-validation (CV) as in \cite{roth2015deeporgan}.   
%%%%%%%%%%%%%%%%%%%%%%%%%%%
As we will see next, candidate generation can be done even more efficiently by using the same HNN architectures, which are based on convolutional neural networks. The technical details of HNNs were described in Sec. \ref{sec:segHNN}. 
%%%%%%%%%%%%%%%%%%%%%%%%%%%%%%%%%%%%%%%%%%%%%%%%%%%%%%%%%%%%%
\subsubsection{Multi-view Aggregated HNNs}\label{sec:localizationHNN}
\label{sec:region_candidates} 
\begin{figure}[htb]%[htb]
\centering	\resizebox{\figscale\columnwidth}{!}{\includegraphics{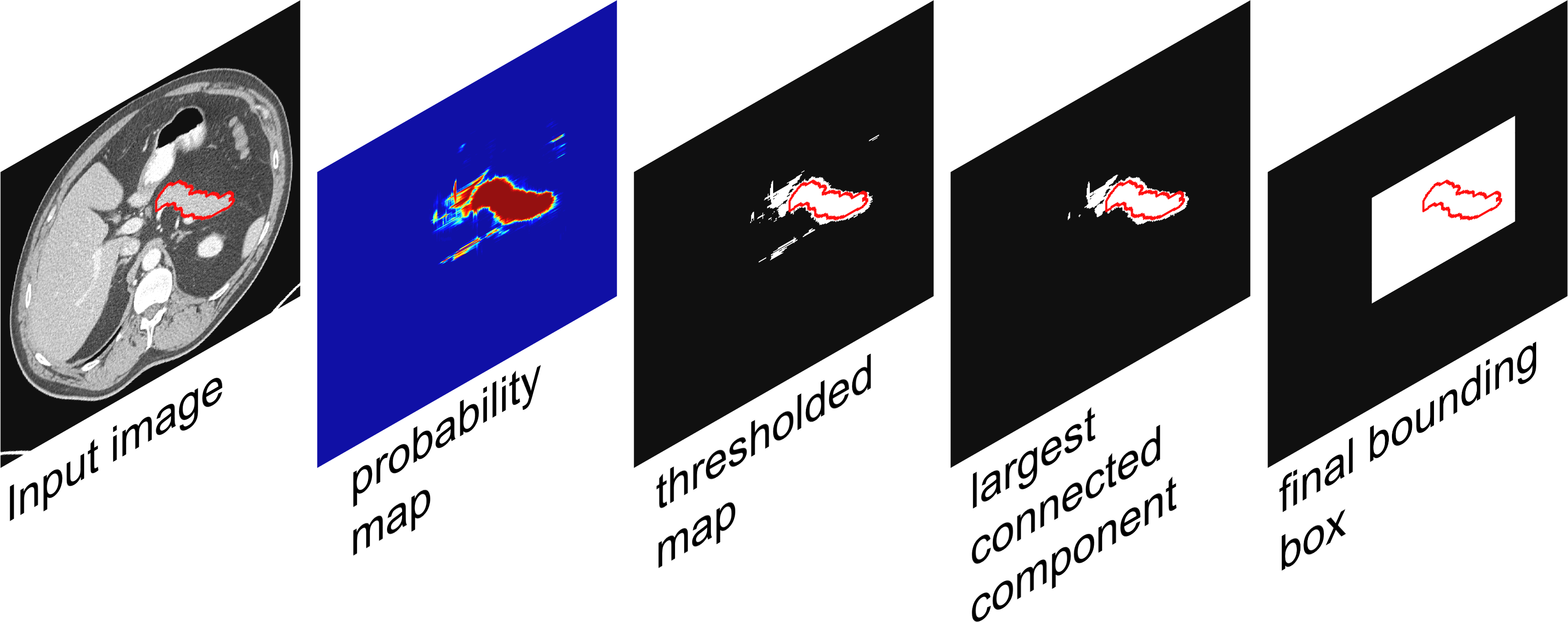}}
	\caption{\small Candidate bounding box region generation pipeline (left to right). Gold standard pancreas in red.}
	\label{fig:bb_pipeline}
\end{figure}

\begin{figure}[htb]%[htb]
\centering	\resizebox{\figscale\columnwidth}{!}{\includegraphics{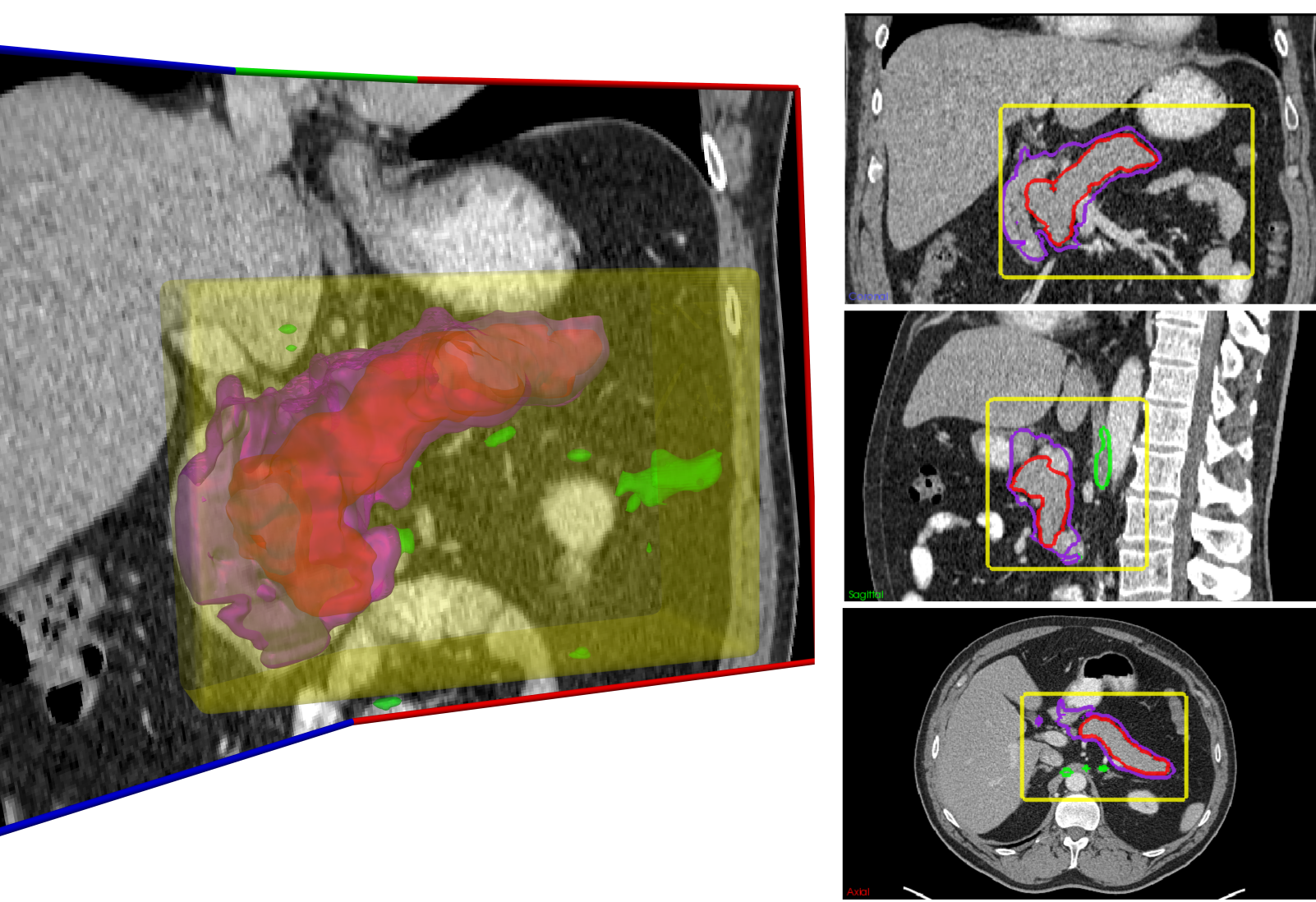}}
	\caption{\small Candidate bounding box region generation. Gold standard pancreas in red, blobs of $\geq0.5$ probabilities in green, the selected largest 3D connected component in purple, the resulting candidate bounding box in yellow.} %(best viewed in color).}
	\label{fig:largest_obj_bb}
\end{figure}

Alternatively to the candidate region generation process described in Sec. \ref{sec:region_candidates_RF} that uses hybrid deep and non-deep learning techniques, we employ \textbf{HNN-I} (interior, see Sec. \ref{sec:segHNN}) as a building block for pancreas localization, inspired by the effectiveness of HNN being able to capture the complex pancreas appearance in CT images \cite{roth2016spatial}. This enables us to drastically discard large negative volumes of the CT scan, while operating \textbf{HNN-I} on a conservative probability threshold of $>=$0.5 that retains high sensitivity/recall ($>$99\%). The constant balancing weight on $\beta$ during training \textbf{HNN-I} is critical in this step since the high majority of CT slices have empty pancreas appearance and are indeed included for effective training of \textbf{HNN-I} models, in order to successfully suppress the pancreas probability values from appearing in background. 
%%%%%%%%%%%%%%%%%%%%%%%%%%%
Furthermore, we perform a largest connected-component analysis to remove outlier ``blobs'' of high probabilities. To get rid of small incorrect connections between high-probability blobs, we first perform an erosion step with radius of 1 voxel, and then select the largest connected-component, and subsequently dilate the region again (Fig. \ref{fig:bb_pipeline}). 
%%%%%%%%%%%%%%%%%%%%%%%%%%%
\textbf{HNN-I} models are trained in axial, coronal, and sagittal planes in order to make use of the multi-view representation of 3D image context. Empirically, we found a max-pooling operation across the 3D models to give the highest sensitivity/recall while still being sufficient to reject the vast amount of non-pancreas from the CT images (see Table \ref{tab:candidate_bb}). One illustrative example is demonstrated in Fig.~\ref{fig:largest_obj_bb}.
%%%%%%%%%%%%%%%%%%%%%%%%%%%%%%%%%%%%%%%%%%%%%%
This initial candidate generation is sufficient to extract bounding box regions that completely surround the pancreases with nearly 100\% recall. All candidate regions are computed during the testing phase of cross-validation (CV) with the same split as in \cite{roth2015deeporgan}. 
%%%%%%%%%%%%%%%%%%%%%%%%%%%%%%%%%%%%%%%%%%%%%%
Note that this candidate region proposal is a crucial step for further processing. It removes ``easy'' non-pancreas tissue from further analysis and allows \textbf{HNN-I} and \textbf{HNN-B} to focus on  the more difficult distinction of pancreas versus its surrounding tissue. The fact that we can use exactly the same HNN model architecture for both stages though is noteworthy.
%%%%%%%%%%%%%%%%%%%%%%%%%%%%%%%%%%%%%%%%%%%%%%%%%%%%%%%%%%%%%
%%%%%%%%%%%%%%%%%%%%%%%%%%%%%%%%%%%%%%%%%%%%%%%%%%%%%%%%%%%%%
%%%%%%%%%%%%%%%%%%%%%%%%%%%%%%%%%%%%%%%%%%%%%%%%%%%%%%%%%%%%%
%%%%%%%%%%%%%%%%%%%%%%%%%%%%%%%%%%%%%%%%%%%%%%%%%%%%%%%%%%%%%
%%%%%%%%%%%%%%%%%%%%%%%%%%%%%%%%%%%%%%%%%%%%%%%%%%%%%%%%%%%%%
\subsection{Pancreas Segmentation} \label{sec:segmentation}
%%%%%%%%%%%%%%%%%%%%%%%%%%%
With pancreas localized, the next step is to produce a reliable segmentation. Our segmentation pipeline consists of three steps. We first use HNN probability maps to generate mid-level boundary and interior cues. These are then used to produce superpixels, which are then aggregated together into a final segmentation using RF classification.
%%%%%%%%%%%%%%%%%%%%%%%%%%%
\subsubsection{Combining Mid-level Cues via HNNs} We now show that organ segmentation can benefit from multiple mid-level cues, like organ interior and boundary predictions. We investigate deep-learning based approaches to independently learn the pancreas' interior and boundary mid-level cues. Combining both cues via learned spatial aggregation can elevate the overall performance of this semantic segmentation system. Organ boundaries are a major mid-level cue for defining and delineating the anatomy of interest. It could prove to be essential for accurate semantic segmentation of an organ.
%%%%%%%%%%%%%%%%%%%%%%%%%%%%%%%%%%%%%%%%%%%%%%%%%%%%%%%%%%%%%
\subsubsection{Learning Organ-specific Segmentation Proposals}\label{sec:segPro} 
\begin{figure*}[htb]%[htb]
 \centering	
    \def\svgwidth{0.85\textwidth}
    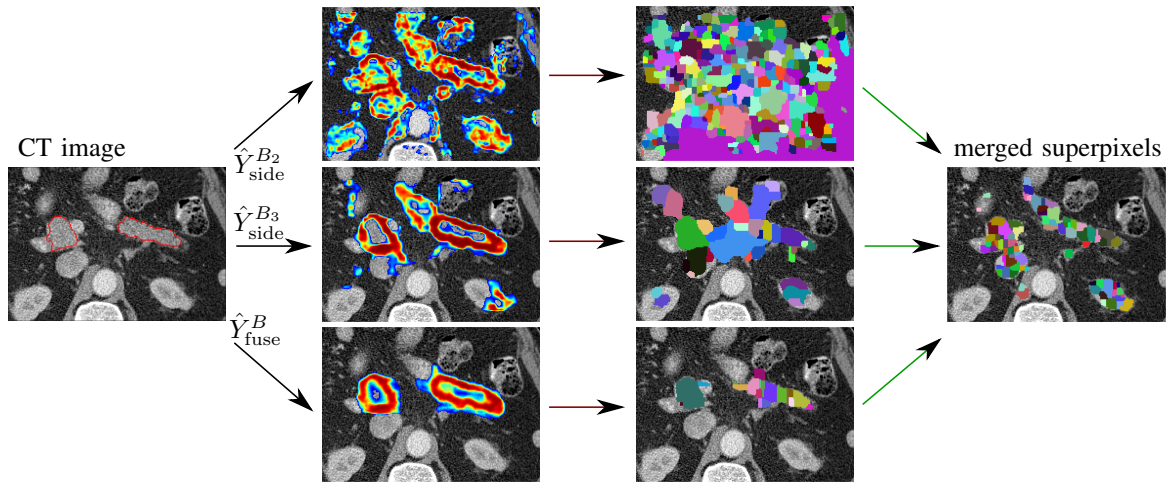    
	\caption{\small Multiscale combinatorial grouping (MCG) \cite{pont-tuset2015mcg} on three different scales of learned boundary predication maps from \textbf{HNN-B}: $\hat{Y}^{B_2}_{\mathrm{side}}$, $\hat{Y}^{B_3}_{\mathrm{side}}$, and $\hat{Y}^{B}_{\mathrm{fuse}}$ using the original CT image on far left as input (with ground truth delineation of pancreas in red). MCG computes superpixels at each scale and produces a set of merged superpixel-based object proposals. We only visualize the boundary probabilities whose values are greater than $.10$ (Figure reproduced from \cite{roth2016spatial}).}
	\label{fig:mcg_superpixels}
\end{figure*}

Multiscale combinatorial grouping (MCG) \cite{pont-tuset2015mcg} is one of the state-of-the-art methods for generating segmentation object proposals in computer vision. We utilize this approach, and publicly available code\footnote{\scriptsize\url{https://github.com/jponttuset/mcg}.}, to generate organ-specific superpixels based on the learned boundary predication maps \textbf{HNN-B}. Superpixels are extracted via continuous oriented watershed transform at three different scales, denoted $(\hat{Y}_\mathrm{side}^{B_2}, \hat{Y}_\mathrm{side}^{B_3} , \hat{Y}_\mathrm{fuse}^{B})$, supervisedly learned by \textbf{HNN-B}. This allows the computation of a hierarchy of superpixel partitions at each scale, and merges superpixels across scales, thereby efficiently exploring their combinatorial space \cite{pont-tuset2015mcg}. This, then, allows MCG to group the merged superpixels toward object proposals. We find that the first two levels of object MCG proposals are sufficient to achieve $\sim88\%$ DSC (see Table \ref{tab:results} and Fig. \ref{fig:mcg_superpixels}), with the optimally computed superpixel labels using their spatial overlapping ratios against the segmentation ground truth map. All \textit{merged} superpixels $\mathcal{S}$ from the first two levels are used for the subsequent spatial aggregation step. Note that \textbf{HNN-B} can only be trained using axial slices where the manual annotation was performed. Pancreas boundary maps in coronal and sagittal views can display strong artifacts.
%%%%%%%%%%%%%%%%%%%%%%%%%%%
%%%%%%%%%%%%%%%%%%%%%%%%%%%%%%%%%%%%%%%%%%%%%%%%%%%%%%%%%%%%%
\subsubsection{Spatial Aggregation with Random Forest}\label{sec:aggregation}
%%%%%%%%%%%%%%%%%%%%%%%%%%%%%%%%%%%%%%%%%%%%%%%%%%%%%%%%%%%%%
We use the superpixel set $\mathcal{S}$ generated previously to extract features for spatial aggregation via random forest classification\footnote{\scriptsize Using MATLAB's TreeBagger() class.}. Within any superpixel $s \in \mathcal{S}$ we compute simple statistics including the 1st-4th order moments, and 8 percentiles $[20\%, 30\%, \dots, 90\%]$ on the CT intensities, and a per-pixel element-wise pooling function of multi-view \textbf{HNN-I}s and \textbf{HNN-B}. Additionally, we compute the mean $x$, $y$, and $z$ coordinates normalized by the range of the 3D candidate region (Sec. \ref{sec:region_candidates}). This results in 39 features describing each superpixel and are used to train a RF classifier on the training positive or negative superpixels at each round of 4-fold CV. %An "optimal" set of superpixels is then calculated that achieve the highest DSC in all cases to decide whether a given superpixel is a \textit{positive} or \textit{negative} training example. 
Empirically, we find 50 trees to be sufficient to model our feature set. A final 3D pancreas segmentation is simply obtained by stacking each slice prediction back into the original CT volume space. No further post-processing is employed. %and spatial aggregation of \textbf{HNN-I} and \textbf{HNN-B} maps for superpixel classification is already of high quality. 
This complete pancreas segmentation model is denoted as \textbf{HNN-RF}. 
%%%%%%%%%%%%%%%%%%%%%%%%%%%%%%%%%%%%%%%%%%%%%%%%%%%%%%%%%%%%%
\section{Experimental Results} \label{sec:results}
\subsection{Data} 
\noindent Manual tracings of the pancreas for \Npatients{} contrast-enhanced abdominal CT volumes are provided by a publicly available dataset\footnote{\scriptsize\url{http://dx.doi.org/10.7937/K9/TCIA.2016.tNB1kqBU}.} \cite{roth2015deeporgan}, for the ease of comparison. Our experiments are conducted on random splits of \Ntrain{} patients for training and \Ntest{} for unseen testing, in 4-fold cross-validation throughout in this section, unless otherwise mentioned. %Most previous work \cite{Wang2014Miccai,Chu2013Miccai,wolz2013automated} use the leave-one-patient-out (LOO) protocol which is computationally expensive (e.g., $\sim15$ hours to process one case using a powerful workstation \cite{Wang2014Miccai}) and may not scale up efficiently towards larger patient populations.
%%%%%%%%%%%%%%%%%%%%%%%%%%%%%%%%%%%%%%%%%%%%%%%%%%%%%%%%%%%%%
\subsection{Evaluation} \label{sec:eva}
\noindent We perform extensive quantitative evaluation on different configurations of our method and compare to the previous state-of-the-art work with in-depth analysis. 
\begin{figure}[htb]
  \newcommand\thisfigscale{0.48}
	\centering
	\begin{tabular}{cc}
			\vcenteredhbox{\includegraphics[width=\thisfigscale\columnwidth]{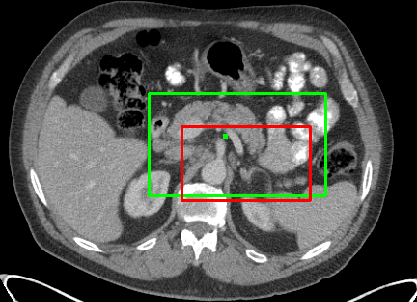}} &
			\vcenteredhbox{\includegraphics[width=0.46\columnwidth]{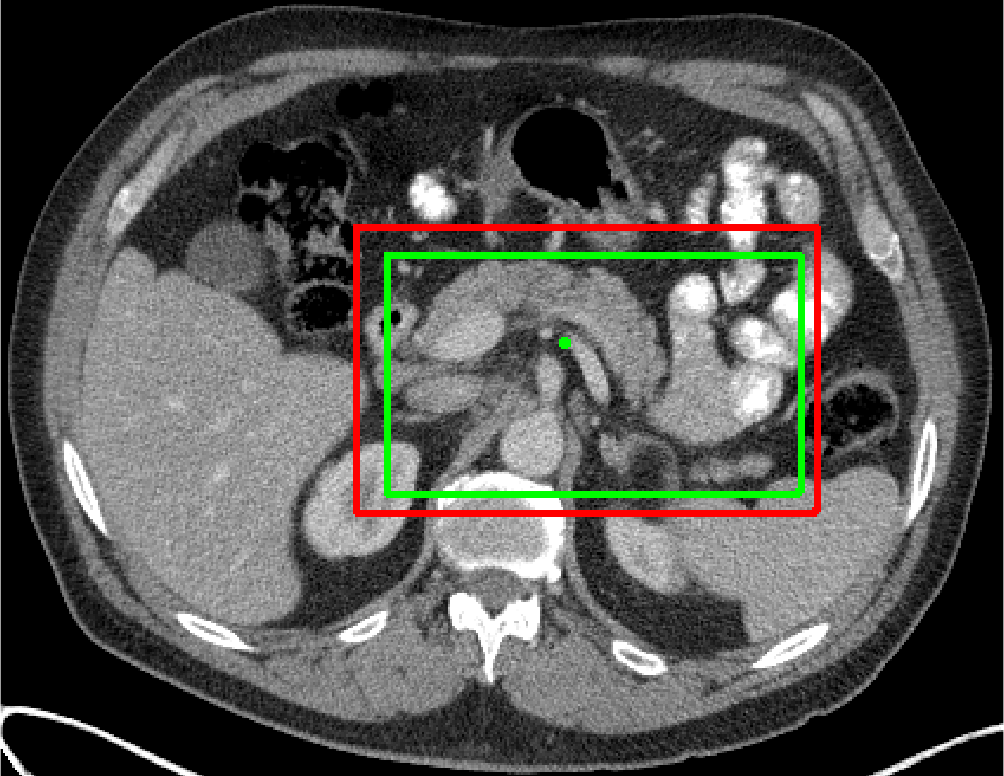}} \\
			(a) \textbf{RF}: Axial & (d) \textbf{HNN-I}: Axial \\
			\vcenteredhbox{\includegraphics[width=\thisfigscale\columnwidth]{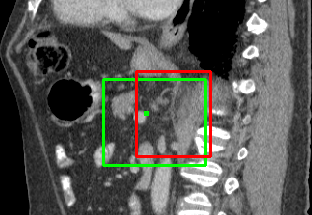}} &	
			\vcenteredhbox{\includegraphics[width=0.46\columnwidth]{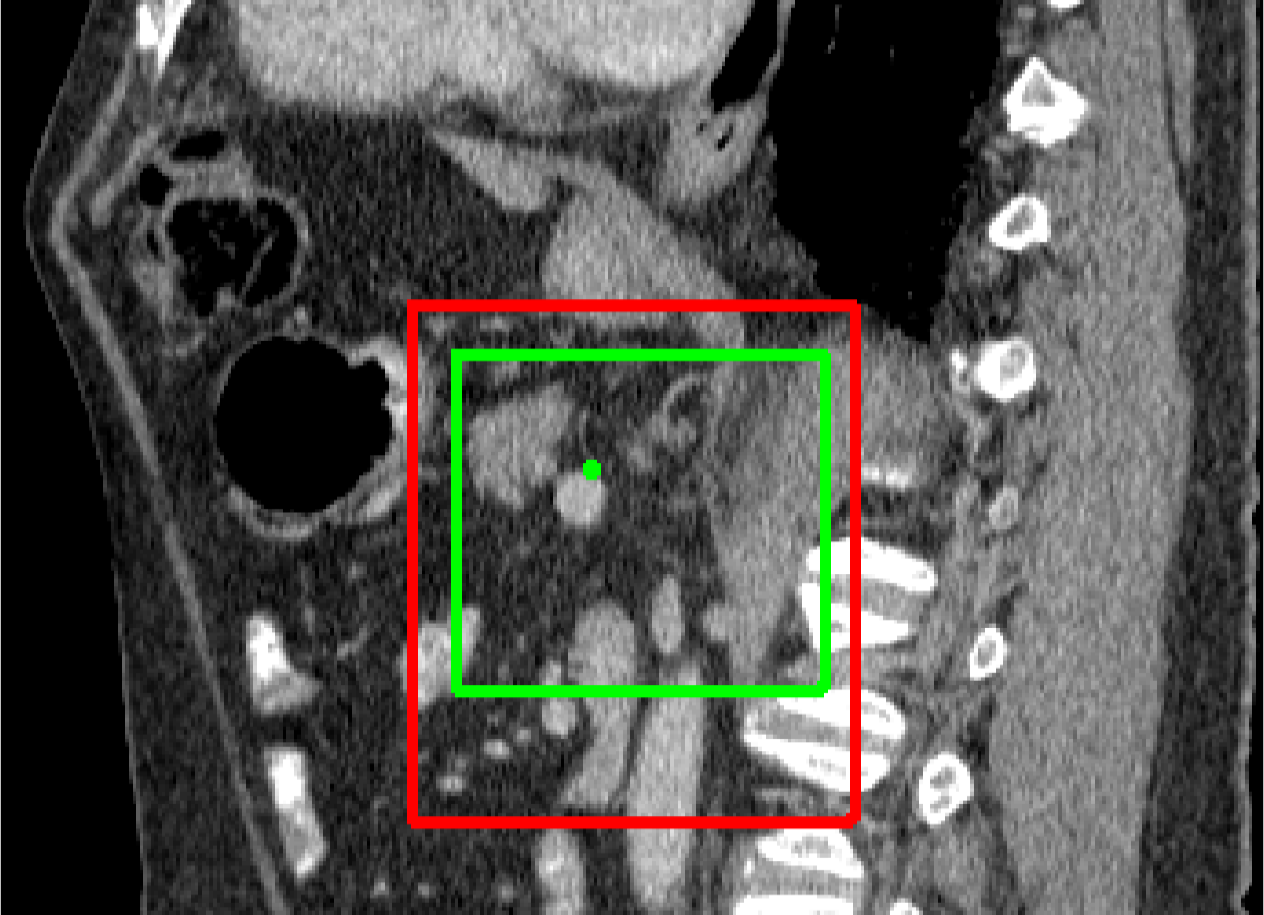}} \\
			(b) \textbf{RF}: Sagittal & (e) \textbf{HNN-I}: Sagittal \\
			\vcenteredhbox{\includegraphics[width=0.47\columnwidth]{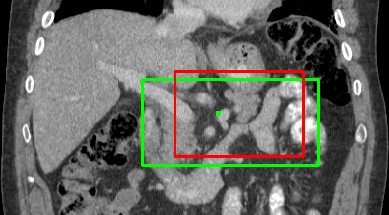}} &
			\vcenteredhbox{\includegraphics[width=\thisfigscale\columnwidth]{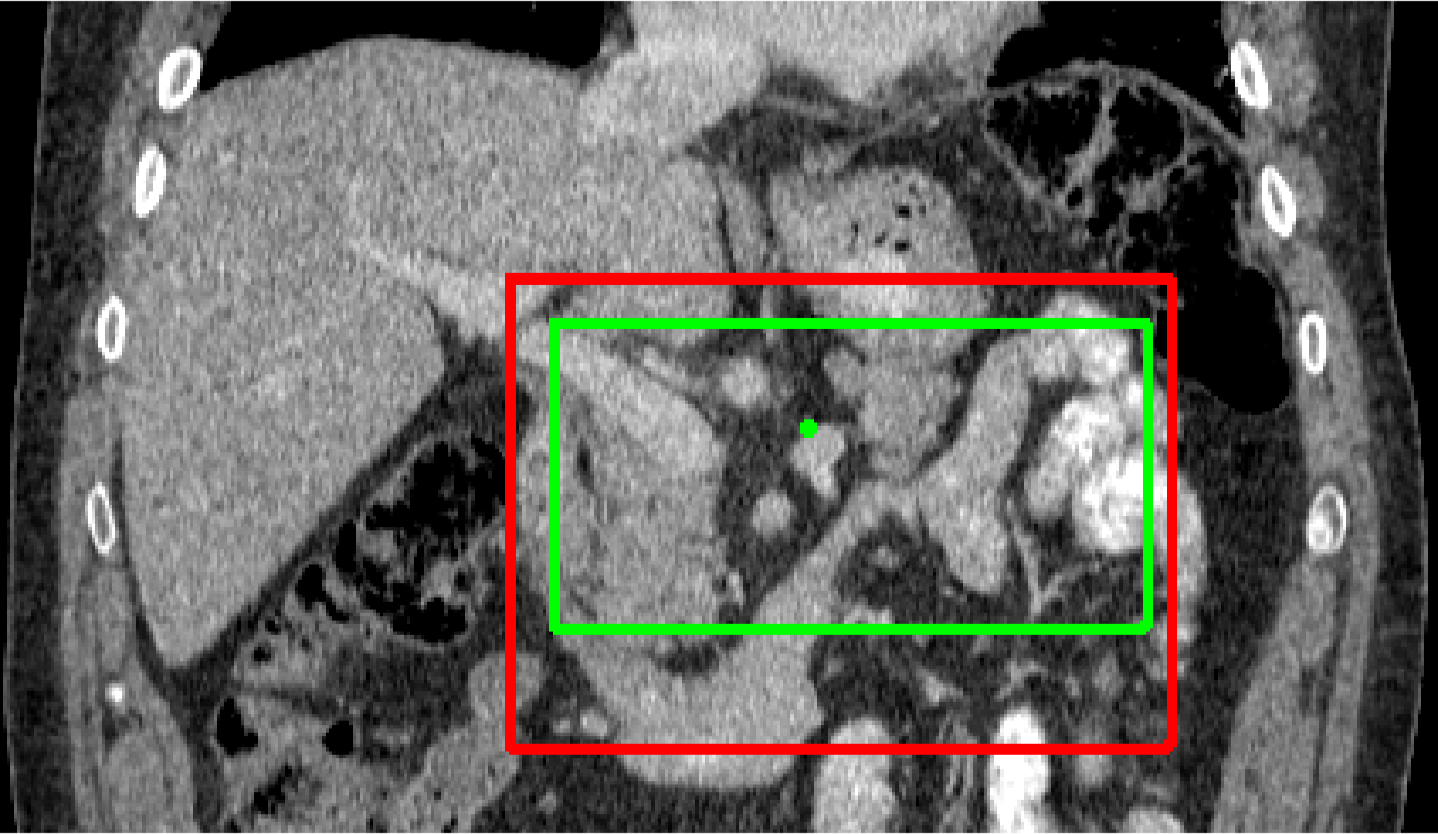}} \\
			(c) \textbf{RF}: Coronal & (f) \textbf{HNN-I}: Coronal \\
	\end{tabular}
	\caption{\small An example for comparison of regression forest (RF, a-c) and \textbf{HNN-I} (d-f) for pancreas localization. Green and red boxes are ground truth and detected bounding boxes respectively. The green dot denotes the ground truth center. This case demonstrates a case in the 90th percentile in RF localization distance and serves as a representative of poorly performing localization. In contrast,  \textbf{HNN-I} includes all of the pancreas with nearly 100\% recall in this case.}
	\label{fig:PancreasForest}
\end{figure}
\subsubsection{Localization}
\begin{figure}[htb]%[htb]
	\centering
	\begin{tabular}{cc}
			\vcenteredhbox{\includegraphics[width=0.46\columnwidth]{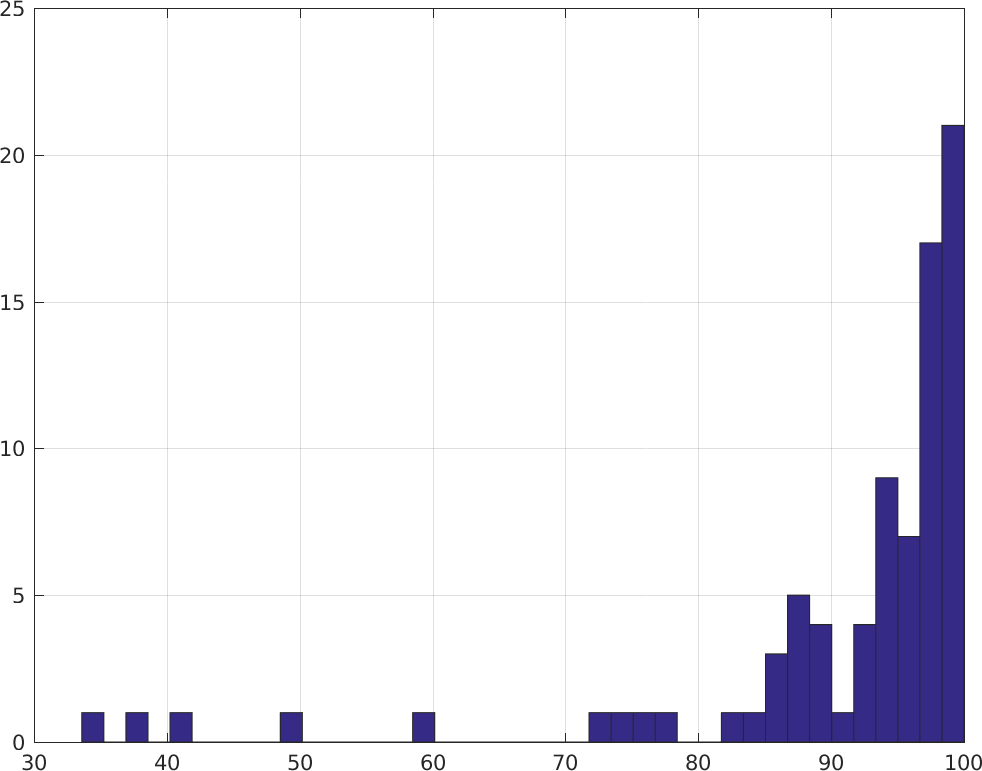}} &
			\vcenteredhbox{\includegraphics[width=0.46\columnwidth]{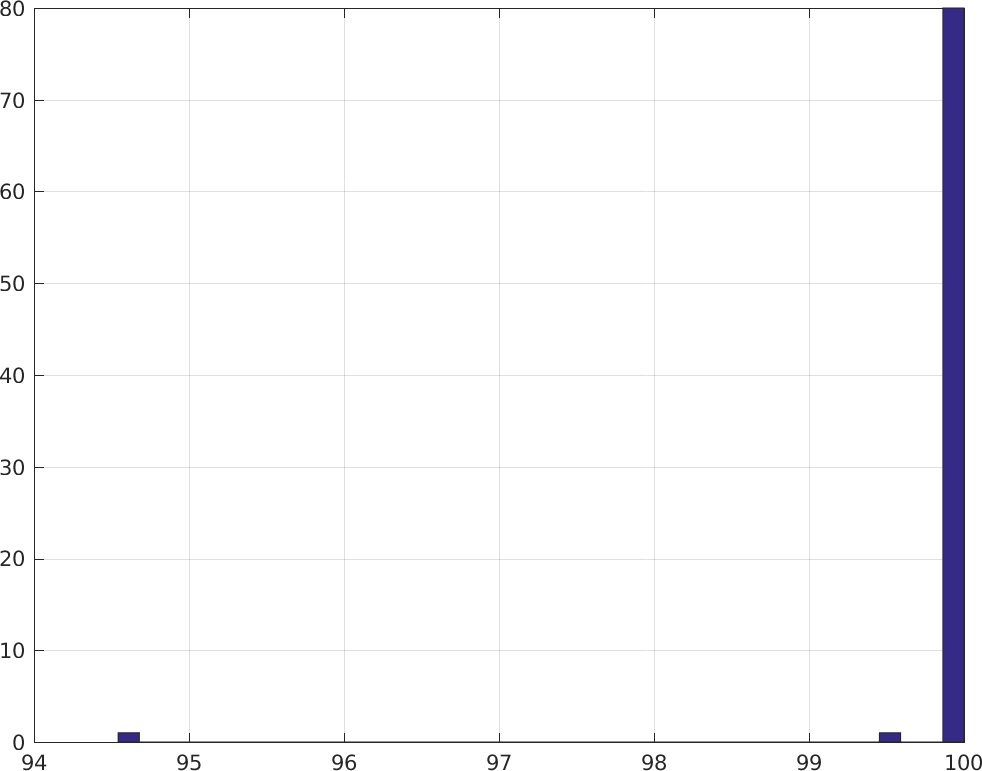}} \\
			(a) regression forest & (b) \textbf{HNN-I}\\
	\end{tabular}
	\caption{\small Histogram plots (Y-Axis) of regression forest based bounding boxes (a) and \textbf{HNN-I}'s generated bounding boxes (b) in recalls (X-axis) covering the ground-truth pancreas masks in 3D. Note that Regression Forest produces 16 out of 82 bounding boxes that lie below 60\% in pixel-to-pixel recall while \textbf{HNN-I} produces 100\% recalls, \hl{except for two cases $\geq$94.54\%}}		
	\label{fig:recall_histograms}
\end{figure}

%%%%%%%%%%%%%%%%%%%%%%%%%%%%%%%%%%%%%%%%%%%%%%    
From our empirical study, the candidate region bounding box generation based on multi-view max-pooled \textbf{HNN-I}s (Sec. \ref{sec:localizationHNN}) or previous hybrid methods (Sec. \ref{sec:region_candidates_RF} \cite{farag2014bottom}) works comparably in terms of addressing the requirement to produce spatially-truncated 3D regions that maximally cover the pancreas in the pixel-to-pixel level and reject as much as possible the background spaces. \hl{An average reduction of absolute volume of 90.36\% (range [80.45\%-96.26\%]) between CT scan and candidate bounding box is achieved during this step, while keeping a mean recall of 99.93\%, ranging [94.54\%-100.00\%]} Table \ref{tbl:PancreasForestPerformance} shows the test performance of pancreas localization and bounding box prediction using regression forests in DSC and average Euclidean distance against the gold standard bounding boxes. As illustrated in Fig. \ref{fig:recall_histograms}, regression forest based localization generates 16 out of 82 bounding boxes that lie below 60\% in the pixel-to-pixel recall against the ground-truth pancreas masks. \hl{Nevertheless we obtain nearly 100\% recall for all scans (except for two cases $\geq$94.54\%) through the multi-view max-pooled \textbf{HNN-I}s. An example of detected pancreas can be seen in Fig.~\ref{fig:PancreasForest}.}
\begin{table}[H]
\centering
\begin{tabular}{c|c|c|c|c|c|c|c}
	\toprule
	\toprule
Metrics & Mean & Std. & 10\% & 50\% & 90\% & Min & Max \\
	\midrule       
Distance (mm) & 14.9 & 9.4 & 6.4 & 11.7 & 29.3 & 2.8 & 48.7 \\
Dice & 0.71 & 0.11 & 0.56 & 0.74 & 0.83 & 0.33 & 0.92 \\
	\bottomrule
	\bottomrule
\end{tabular}
\caption{\small Test performance of pancreas localization and bounding box prediction using regression forests in Dice and average Euclidean distance against the gold standard bounding boxes, in 4-fold cross validation.} 
%\textcolor{red}{Adam: should you also not include RF and HNN results in this table?}
\label{tbl:PancreasForestPerformance}
\end{table}
%%%%%%%%%%%%%%%%%%%%%%%%%%%%%%%%%%%%%%%%%%%%%%%%%%%%%%%%%%%%%
%%%%%%%%%%%%%%%%%%%%%%%%%%%%%%%%%%%%%%%%%%%%%%%%%%%%%%%%%%%%%
%%%%%%%%%%%%%%%%%%%%%%%  
\subsubsection{HNN Spatial Aggregation for Pancreas Segmentation}
%%%%%%%%%%%%%%%%%%%%%%%  
The interior HNN models trained on the axial (AX), coronal (CO) or sagittal (SA) CT images in Sec. \ref{sec:localizationHNN} can be straightforwardly used to generate pancreas segmentation masks. We exploit different spatial aggregation or pooling functions on the AX, CO, and SA viewed \textbf{HNN-I} probability maps, denoted as \textbf{AX}, \textbf{CO}, \textbf{SA} (any single view \textbf{HNN-I} probability map simply used); \textbf{mean(AX,CO)}, \textbf{mean(AX,SA)}, \textbf{mean(CO,SA)} and \textbf{mean(AX,CO,SA)} (element-wise mean of two or three view \textbf{HNN-I} probability maps); \textbf{max(AX,CO,SA)} (element-wise maximum of three view \textbf{HNN-I} probability maps); and finally \textbf{meanmax(AX,CO,SA)} (element-wise mean of the maximal two scores from three view \textbf{HNN-I} probability maps). After the optimal thresholding calibrated using the training folds on these pooled \textbf{HNN-I} maps, the resulting binary segmentation masks are further refined by 3D connected component process and simple morphological operations (as in Sec. \ref{sec:localizationHNN}). Table \ref{tab:candidate_bb} demonstrates the DSC pancreas segmentation accuracy performance by investigating different spatial aggregation functions. We observe that the element-wise multi-view (mean or max) pooling operations on \textbf{HNN-I} probabilities maps generally outperform their single view counterparts. \textbf{max(AX,CO,SA)} performs slightly better than \textbf{mean(AX,CO,SA)}. The configuration of \textbf{meanmax(AX,CO,SA)} produces the most superior performance in mean DSC which may behave as a robust fusion function by rejecting the smallest probability value and averaging the remained two \textbf{HNN-I} scores per pixel location. 
%%%%%%%%%%%%%%%%%%%%%%%%%%%%%%%%%%%%%%%%%%%%%%    
\begin{table}[htb]
\centering
\caption{\small  \textbf{Four-fold cross-validation}: DSC [\%] pancreas segmentation performance of various spatial aggregation functions on AX, CO, and SA viewed \textbf{HNN-I} probability maps in the candidate region generation stage (the best results in \textbf{bold}).}    
    \begin{tabular}{l|c|c|c|c}
	\toprule
	\toprule
\textbf{DSC} &	\textbf{Mean} & \textbf{Std} & \textbf{Min} & \textbf{Max} \\
	\midrule       
\textbf{AX} &	73.46 &	11.63 &	1.88 &	85.97 \\
\textbf{CO} &	70.19 &	\textbf{9.81} &	\textbf{39.72} &	83.84 \\
\textbf{SA} &	72.42 &	11.26 &	14.00 &	84.92 \\
\textbf{mean(AX,CO)} &	74.65 &	11.21 &	5.08 &	86.87 \\ 
\textbf{mean(AX,SA)} &	75.08 &	12.29 &	2.31 &	86.97 \\
\textbf{mean(CO,SA)} &     73.70 &       11.40 &      18.96 & 86.64 \\
\textbf{mean(AX,CO,SA)} &	75.07 &	12.08 &	4.26 &	87.19 \\
\textbf{max(AX,CO,SA)} &	75.67 &	10.32 &	16.11 &	87.65 \\
\textbf{meanmax(AX,CO,SA)} & \textbf{76.79} & 11.07  & 8.97 & \textbf{88.03} \\
    	\bottomrule
	\bottomrule
    \end{tabular}%
		 \label{tab:candidate_bb}
\end{table}
%%%%%%%%%%%%%%%%%%%%%%%%%%%%%%%%%%%%%%%%%%%%%%    
\begin{table}[htb]
\centering
\caption{\small  \textbf{Four-fold cross-validation}: DSC [\%] pancreas segmentation performance of various spatial aggregation functions on AX, CO, and SA viewed \textbf{HNN-I} probability maps in the second cascaded stage (the best results in \textbf{bold}).}
\begin{tabular}{l|c|c|c|c}
\toprule
\toprule
\textbf{DSC} &	\textbf{Mean} & \textbf{Std} & \textbf{Min} & \textbf{Max} \\
\midrule       
\textbf{AX} 					  & 78.99 & 7.70  & 44.25 & 88.69\\
\textbf{CO} 					  & 76.16 & 8.67  & 45.29 & 88.11\\
\textbf{SA} 					  & 76.53 & 9.35  & 40.60 & 88.34\\
\textbf{mean(AX,CO)} 				  & 79.02 & 7.96  & 42.64 & 88.82\\
\textbf{mean(AX,SA)}				  & 79.29 & 8.21  & 42.32 & 89.38\\ 
\textbf{mean(CO,SA)}                               & 77.61 & 8.92  & 44.14 & 89.11\\
\textbf{mean(AX,CO,SA)}                       & 80.40 & 7.30  & 45.18 & 89.11\\
\textbf{max(AX,CO,SA)}               & 80.55 & \textbf{6.89}  & \textbf{45.66} & 89.92\\
\textbf{meanmax(AX,CO,SA)} & \textbf{81.14} & 7.30  & 44.69 & \textbf{89.98}\\
    	\bottomrule
	\bottomrule
    \end{tabular}%
		\label{tab:MV_aggregation}
\end{table}
%%%%%%%%%%%%%%%%%%%%%%%  
After the pancreas localization stage, we train a new set of multi-view \textbf{HNN-I}s with the spatially truncated scales and extents. This serves a desirable ``Zoom Better to See Clearer'' effect for deep neural network segmentation models \cite{Xia2016Zoom} where cascaded \textbf{HNN-I}s only focus on discriminating or parsing the remained organ candidate regions. Similarly, DSC [\%] pancreas segmentation accuracy results of various spatial aggregation or pooling functions on AX, CO, and SA viewed \textbf{HNN-I} probability maps (trained in the second cascaded stage) are shown in Table \ref{tab:MV_aggregation}. We find consistent empirical observations as above when comparing multi-view HNN pooling operations. The \textbf{meanmax(AX,CO,SA)} operation again reports the best mean DSC performance at 81.14\% which is increased considerably from 76.79\% in Table \ref{tab:candidate_bb}. We denote this system configuration as $\mathbf{HNN}_\mathrm{meanmax}$. This result validates our two staged pancreas segmentation framework of proposing candidate region generation for organ localization followed by ``Zoomed''  deep HNN models to refine segmentation.
%%%%%%%%%%%%%%%%%%%%%%%%%%%
Table \ref{tab:results} shows the improvement from the \hl{\textbf{meanmax}-pooled} \textbf{HNN-I}s (i.e., $\mathbf{HNN}_\mathrm{meanmax}$) to the \textbf{HNN-RF} based spatial aggregation, using DSC and average minimum surface-to-surface distance (AVGDIST). The average DSC is increased from 81.14\% to 81.27\%, However, this improvement is not statistically significantly with $p>0.05$ using Wilcoxon signed rank test. In contrast, using dense CRF (DCRF) optimization \cite{chen2014semantic} (with \textbf{HNN-I} as the unary term and the pairwise term depending on the CT values) as a means of introducing spatial consistency does not improve upon \textbf{HNN-I} noticeably as shown in \cite{roth2016spatial}). Comparing to the performance of previous state-of-the-art methods \cite{roth2015deeporgan, roth2016spatial} at mean DSC scores of 71.4\% and 78.01\% respectively, both variants of $\mathbf{HNN}_\mathrm{meanmax}$ and \textbf{HNN-RF} demonstrate superior quantitative segmentation accuracy in DSC and AVGDIST metrics. We have the following two observations. 1, The main performance gain from \cite{roth2016spatial} (similar to $HNN_{AX}$ in Table \ref{tab:MV_aggregation}) is found by the multi-view aggregated HNN pancreas segmentation probability maps (e.g., $\mathbf{HNN}_\mathrm{meanmax}$), which also serve in \textbf{HNN-RF}. 2, The new candidate region bounding box generation method (Sec. \ref{sec:localizationHNN}) works comparably to the hybrid technique (Sec. \ref{sec:region_candidates_RF} \cite{farag2014bottom,roth2015deeporgan,roth2016spatial}) based on our empirical evaluation. However the proposed pancreas localization via multi-view max-pooled HNNs greatly simplified our overall pancreas segmentation system which may also help the generality and reproducibility. The variant of $\mathbf{HNN}_\mathrm{meanmax}$ produces competitive segmentation accuracy but merely involves evaluating two sets of multi-view \textbf{HNN-I}s at two spatial scales: whole CT slices or truncated bounding boxes. There is no need to compute any hand-crafted image features \cite{farag2014bottom} or train other external machine learning classifiers. As shown in Fig. \ref{fig:recall_histograms}, the conventional organ localization framework using regression forest \cite{criminisi2013regression,laybirkbeck2013} does not address well the purpose of candidate region generation for segmentation where extremely high pixel-to-pixel recall is required since it is mainly designed for organ detection.
%%%%%%%%%%%%%%%%%%%%%%%%%%%
In Table \ref{tab:results2}, the quantitative pancreas segmentation performance of two method variants, $\mathbf{HNN}_\mathrm{meanmax}$, \textbf{HNN-RF} spatial aggregation, are evaluated using four metrics of DSC (\%), Jaccard Index (\%) \cite{Levandowsky1971}, Hausdorff distance (HDRFDST [mm]) \cite{Rockafellar2005} and AVGDIST [mm]. Note that there is no statistical significance when comparing the performance of two variants in three measures of DSC, JACARD, and AVGDIST, except for HDRFDIST with $p<0.001$ under Wilcoxon signed rank test. Since Hausdorff distance represents the maximum deviation between two point sets or surfaces, this observation indicates that \textbf{HNN-RF} may be more robust than $\mathbf{HNN}_\mathrm{meanmax}$ in the worst case scenario.
%%%%%%%%%%%%%%%%%%%%%%%%%%%%%%%%%%%%%%%%%%%%%%%%%%%%%%%%%%%%%

\begin{figure*}[htb]%[htb]
\centering	\resizebox{0.85\linewidth}{!}{\includegraphics{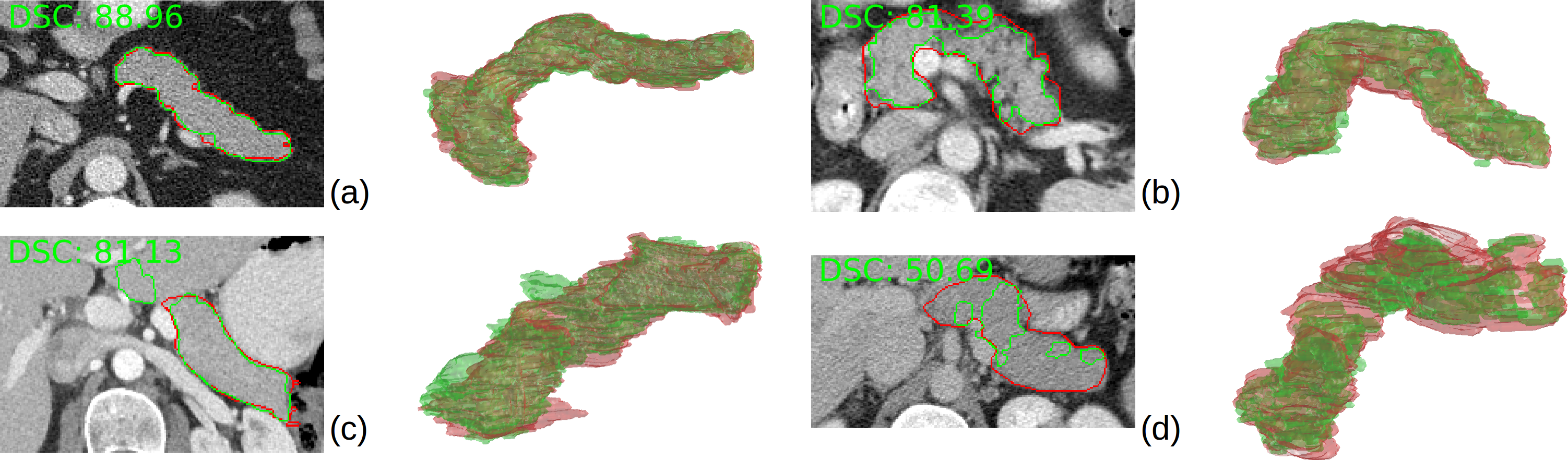}}
	\caption{\small Examples of our \textbf{HNN-RF} pancreas segmentation results (green) comparing with the ground-truth annotation (red). The best performing case (a), two cases with DSC scores close to the data set mean (b,c) and the worst case are shown (d).}
	\label{fig:axial_examples}
\end{figure*}

%%%%%%%%%%%%%%%%%%%%%%%%%%%%%%%%%%%%%%%%%%%%%%%%%%%%%%%%%%%%%
Pancreas segmentation on illustrative patient cases are shown in Fig. \ref{fig:axial_examples}. Furthermore, we applied our trained \textbf{HNN-I} model on a different CT data set\footnote{\scriptsize 30 training data sets at \url{https://www.synapse.org/\#!Synapse:syn3193805/wiki/217789}.} with 30 patients, and achieve a mean DSC of 62.26\% without any re-training on the new data cases, but if we average the outputs of our 4 \textbf{HNN-I} models from cross-validation, we achieve 65.66\% DSC. This demonstrates that \textbf{HNN-I} may be highly generalizable in cross-dataset evaluation. Performance on that dataset will likely improve with further fine-tuning. Last, we collected an additional dataset of 19 unseen CT scans using the same patient data protocol \cite{roth2015deeporgan,roth2016spatial}. Here, $\mathbf{HNN}_\mathrm{meanmax}$ achieves a mean DSC of 81.2\%.  
%%%%%%%%%%%%%%%%%%%%%%%%%%%%%%%%%%%%%%%%%%%%%%%%%%%%%%%%%%%%%
\begin{table}[htb]
%\small
%\footnotesize
%\scriptsize
%\tiny
\centering
 \caption{\small \textbf{Four-fold cross-validation}: The DSC [\%] and average surface-to-surface minimum distance (AVGDIST [mm]) performance of \cite{roth2015deeporgan}, \cite{roth2016spatial}, $\mathbf{HNN}_\mathrm{meanmax}$, \textbf{HNN-RF} spatial aggregation, and optimally achievable superpixel assignments (\textit{italic}). Best performing method in \textbf{bold}.} 
 %%%%%%%%%%%%%%%%%%%%%%%  
 \begin{subtable}[center]{0.95\columnwidth}
  \raggedright
    \begin{tabular}{l|c|c|c|c|c}
    \toprule
		\toprule
    \textbf{DSC} & \cite{roth2015deeporgan}  & \cite{roth2016spatial}& $\mathbf{HNN}_\mathrm{meanmax}$ & \textbf{HNN-RF} & \textit{\textbf{Opt.}} \\
		\midrule    
\textbf{Mean}	&71.42 & 78.01& 81.14                 &\textbf{81.27}  	& \textit{87.67}   \\
\textbf{Std}      &10.11 & 8.20  	      & 7.30                   &\textbf{6.27} & \textit{ 2.21}  \\
\textbf{Min}	&23.99 & 34.11 	   &	44.69                  &\textbf{50.69} & \textit{81.59}\\
\textbf{Max}	&86.29 & 88.65  	   &	\textbf{89.98} &88.96 & \textit{ 91.71} \\
    \bottomrule
		\bottomrule
    \end{tabular}%
 \end{subtable}%
%%%%%%%%%%%%%%%%%%%%%%%    
\vskip 8pt  
 \begin{subtable}[center]{0.95\columnwidth}
 \raggedright
    \begin{tabular}{l|c|c|c|c|c}
    \toprule
		\toprule
    \textbf{AVGDIST} & \cite{roth2015deeporgan} & \cite{roth2016spatial} & $\mathbf{HNN}_\mathrm{meanmax}$ & \textbf{HNN-RF} & \textit{\textbf{Opt.}}\\
		\midrule    
\textbf{Mean}	&1.53	&0.60    	&0.43 &\textbf{0.42}&\textit{0.16}\\
\textbf{Std}      &1.60	&0.55		&0.32 &\textbf{0.31}&\textit{0.04}\\
\textbf{Min}	&0.20	&0.15	&\textbf{0.12} &0.14&\textit{0.10}	\\
\textbf{Max}	&10.32	&4.37		&\textbf{1.88} &2.26&\textit{0.26}\\
    \bottomrule
		\bottomrule
    \end{tabular}%
 \end{subtable}
%%%%%%%%%%%%%%%%%%%%%%%    
\label{tab:results}%
\end{table}%
%%%%%%%%%%%%%%%%%%%%%%%%%%%%%%%%%%%%%%%%%%%%%%%%%%%%%%%%%%%%%
\begin{figure*}[htb]%[htb]
\centering	\resizebox{0.85\textwidth}{!}{\includegraphics{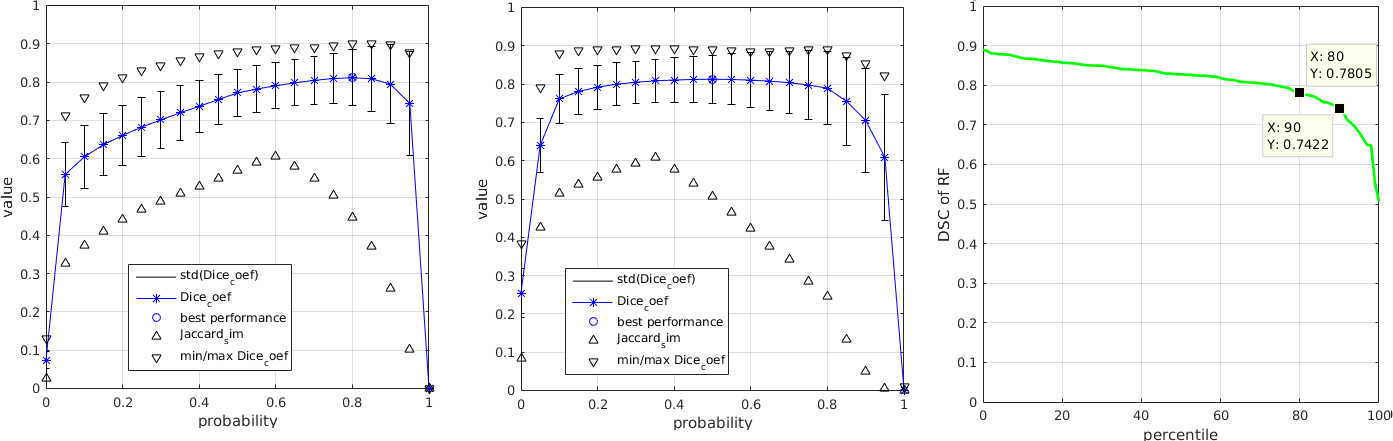}}
	\caption{\small Average DSC performance as a function of pancreas probability using $\mathbf{HNN}_\mathrm{meanmax}$ (left) and spatial aggregation via \textbf{RF} (middle) for comparison. Note that the DSC performance remains much more stable after \textbf{RF} aggregation with respect to the probability threshold. The percentage of total cases that lie  above a certain DSC with \textbf{RF} are shown (right): 80\% of the cases have a DSC of 78.05\%, and 90\% of the cases have a DSC of 74.22\% and higher.}
	\label{fig:avg_dsc}
\end{figure*}
%%%%%%%%%%%%%%%%%%%%%%%%%%%%%%%%%%%%%%%%%%%%%%%%%%%%%%%%%%%%%
\begin{table*}[htb]
\begin{center}
\caption{\small \textbf{Four-fold cross-validation}: The quantitative pancreas segmentation performance results of our two method variants, $\mathbf{HNN}_\mathrm{meanmax}$, \textbf{HNN-RF} spatial aggregation, in four metrics of DSC (\%), Jaccard Index (\%), Hausdorff distance (HDRFDST [mm]), and AVGDIST [mm]. Best performing methods are shown in \textbf{bold}. Note that there is no statistical significance when comparing the performance by two variants in three measures of DSC, JACARD, and AVGDIST, except for HDRFDIST with $p<0.001$ (Wilcoxon Signed Rank Test). This indicates that \textbf{HNN-RF} may be more robust than $\mathbf{HNN}_\mathrm{meanmax}$ in the worst case scenario.}
\label{tab:results2}
%%%%%%%%%%%%%%%%%%%%%%%  
\begin{tabular}{l|c|c|c|c|c|c|c|c}
    \toprule
		\toprule
    ~ & \multicolumn{2}{|c|}{DSC} & \multicolumn{2}{|c|}{Jaccard} & \multicolumn{2}{|c|}{HDRFDST}  & \multicolumn{2}{|c|}{AVGDIST} \\
		\midrule 
		~ & $\mathbf{HNN}_\mathrm{meanmax}$ & \textbf{HNN-RF} & $\mathbf{HNN}_\mathrm{meanmax}$ & \textbf{HNN-RF} & $\mathbf{HNN}_\mathrm{meanmax}$ & \textbf{HNN-RF} & $\mathbf{HNN}_\mathrm{meanmax}$ & \textbf{HNN-RF} \\	
		\midrule 	
\textbf{Mean}	    &81.14 & \textbf{81.27}   &68.82 &\textbf{68.87}    &22.24 & \textbf{17.71}   &0.43 & \textbf{0.42}   \\
\textbf{Std}      &7.30  & \textbf{6.27}  	&9.27  &\textbf{8.12}     &13.90 & \textbf{10.40}   &0.32 & \textbf{0.31}  \\
\textbf{Median}   &\textbf{82.98} & 82.75		&\textbf{70.92} &70.57    &18.03 & \textbf{14.88}	  &0.32	&	0.32	\\
\textbf{Min}	    &44.69 & \textbf{50.69} 	&28.78 &\textbf{33.95}    &5.83  & \textbf{5.20}	  &\textbf{0.12}		&0.14 \\
\textbf{Max}	    &\textbf{89.98} & 88.96  	&79.52 &\textbf{80.12}    &79.52 & \textbf{69.14}   &\textbf{1.88}		&2.26 \\
    \bottomrule
		\bottomrule
\end{tabular}%
\end{center} 
\end{table*}%
%%%%%%%%%%%%%%%%%%%%%%%%%%%%%%%%%%%%%%%%%%%%%%%%%%%%%%%%%%%%%
%%%%%%%%%%%%%%%%%%%%%%%%%%%%%%%%%%%%%%%%%%%%%%%%%%%%%%%%%%%%%
\section{Discussion \& Conclusion}\label{sec:discussion}
%%%%%%%%%%%%%%%%%%%%%%%%%%%%%%%%%%%%%%%%%%%%%%%%%%%%%%%%%%%%%
\noindent To the best of our knowledge, our result comprises the highest reported average DSC in testing folds under 4-fold CV evaluation metric. Strict comparison to other methods (except for \cite{roth2015deeporgan,roth2016spatial}) is not directly possible due to different datasets utilized. Our holistic segmentation approach with multi-view pooling and spatial aggregation advances the current state-of-the-art quantitative performance to an average DSC of 81.27\% in testing. Previous notable results for CT images range from $\sim$68\% to $\sim$78\% \cite{wolz2013automated,Chu2013Miccai,tong2015discriminative,okada2015abdominal,oda2016regression}, all under the ``leave-one-patient-out'' (LOO) cross-validation scheme. In particular, DSC drops from 68\% (150 patients) to 58\% (50 patients) as reported in \cite{wolz2013automated}. Our methods also perform with the better statistical stability, i.e., comparing 7.3\%  or 6.27\% versus 18.6\% \cite{Wang2014Miccai}, 15.3\% \cite{Chu2013Miccai} in the standard deviation of DSC scores. The minimal DSC values are 44.69\% with $\mathbf{HNN}_\mathrm{meanmax}$ and 50.69\% for \textbf{HNN-RF} whereas \cite{Wang2014Miccai,Chu2013Miccai,wolz2013automated,roth2015deeporgan} all report patient cases with DSC $<$10\%. 
%%%%%%%%%%%%%%%%%%%%%%%%%%%%%%%%%%%%%%%%%%%%%%%%%%%%%%%%%%%%%
%%%%%%%%%%%%%%%%%%%%%%%%%%%
Recent work that explores the direct application of 3D convolutional filters with fully convolutional architectures also shows promise \cite{ronneberger2015unet,Merkow2016Dense}. It has to be established whether 2D or 3D implementations are more suited for certain tasks. There is some evidence that deep networks representations with direct 3D input suffer from the \textit{curse-of-dimensionality} and are more prone to overfitting \cite{roth2016improving,Su2015Multi,Su2016Volumetric}. Volumetric object detection might require more training data and might suffer from scalability issues. However, proper hyper-parameter tuning of the CNN architecture and enough training data (including data augmentation) might help eliminate these problems. In the mean time, spatial aggregation in multiple 2D views (as proposed here) might be a very efficient (and computationally less expensive) way of diminishing the curse-of-dimensionality. Furthermore, using 2D views has the advantage that networks trained on much larger databases of natural images (e.g. \textit{ImageNet, BSDS500}) can be used for fine-tuning to the medical domain. It has been shown that transfer learning is a viable approach when the medical imaging data set size is limited \cite{Shin2016Deep,tajbakhsh2016convolutional}. 3D CNN approaches often adopt padded spatially-local sliding volumes to parse any CT scan, e.g., 96$\times$96$\times$48 \cite{Merkow2016Dense}, 160$\times$160$\times$72 \cite{Dou20163D} or 80$\times$80$\times$80 \cite{Chen2016VoxResNet}, which may cause the segmentation discontinuity or inconsistency at overlapped window boundaries. Ensemble of several neural networks trained with random configuration variations is found to be advantageous comparing a single CNN model in object recognition \cite{simonyan2014very,krizhevsky2012imagenet,Simonyan2014Two}. Our pancreas segmentation method can be indeed considered as ensembles of multiple correlated HNN models but good complementary information gain since they are trained from orthogonal axial, coronal or sagittal CT views.  
%%%%%%%%%%%%%%%%%%%%%%%%%%%%%%%%%%%%%%%%%%%%%%%%%%%%%%%%%%%%%
%%%%%%%%%%%%%%%%%%%%%%%%%%%
In conclusion, we present a holistic deep CNN approach for pancreas localization and segmentation in abdominal CT scans, exploiting multi-view spatial pooling and combining interior and boundary mid-level cues. The robust fusion of $\mathbf{HNN}_\mathrm{meanmax}$ aggregating on interior holistically-nested networks ({\bf HNN-I}) alone already achieve good performance at DSC of 81.14\%$\pm$7.30\% in 4-fold CV. The other method variant \textbf{HNN-RF} incorporates the organ boundary responses from the {\bf HNN-B} model and significantly improves the worst case pancreas segmentation accuracy in Hausdorff distance (p$<$0.001). The highest reported DSCs of 81.27\%$\pm$6.27\% is achieved, at the computational cost of 2$\sim$3 minutes, not hours as in \cite{Wang2014Miccai,Chu2013Miccai,wolz2013automated}. Our deep learning based organ segmentation approach could be generalizable to other segmentation problems with large variations and pathologies, e.g., pathological organs and tumors.
%%%%%%%%%%%%%%%%%%%%%%%%%%%%%%%%%%%%%%%%%%%%%%%%%%%%%%%%%%%%%
%%%%%%%%%%%%%%%%%%%%%%%%%%%%%%%%%%%%%%%%%%%%%%%%%%%%%%%%%%%%%
%\begin{table}
%	\centering
%		\begin{tabular}
%			DSC Use Amal`s paper? Mori paper
%		\end{tabular}
%	\caption{\small Comparison to other methods of pancreas segmentation.}
%	\label{tab:ComparisonToOtherMethodsOfPancreasSegmentation}
%\end{table}
%%%%%%%%%%%%%%%%%%%%%%%%%%%%%%%%%%%%%%%%%%%%%%%%%%%%%%%%%%%%%
%%%%%%%%%%%%%%%%%%%%%%%%%%%%%%%%%%%%%%%%%%%%%%%%%%%%%%%%%%%%%
%\section*{Acknowledgment}
%This work was supported by the Intramural Research Program of National Institutes of Health Clinical Center.
%%%%%%%%%%%%%%%%%%%%%%%%%%%%%%%%%%%%%%%%%%%%%%%%
% BIBLIOGRAPHY
%%%%%%%%%%%%%%%%%%%%%%%%%%%%%%%%%%%%%%%%%%%%%%%%
%\clearpage
%\newpage
\bibliographystyle{ieeetr}  % (1) prints author names abbreviated (like {abbrv}) in the references section and (2) sorts references numerically in citation order (like {unsrt})
\bibliography{references_tmi2016}
%\vfill
\end{document}